\documentclass[10pt,twocolumn,letterpaper]{article}
\makeatletter

\newcommand{\thickhline}{\hline } 
\makeatother
\usepackage{cvpr}
\usepackage{times}
\usepackage{epsfig}
\usepackage{graphicx}
\usepackage{amsmath}
\usepackage{amssymb}

\usepackage[pagebackref=true,breaklinks=true,letterpaper=true,colorlinks,bookmarks=false]{hyperref}

\cvprfinalcopy 

\def\cvprPaperID{2368} 

\ifcvprfinal\fi
\title{AutoAugment:\\ Learning Augmentation Strategies from Data}

\author{Ekin D. Cubuk \,\thanks{Equal contribution.}\;, Barret Zoph\footnotemark[1]\,, Dandelion Man\'e, Vijay Vasudevan, Quoc V. Le \\ Google Brain
}

\begin{document}

\maketitle

\begin{abstract}
Data augmentation is an effective technique for improving the accuracy of modern image classifiers. However, current data augmentation implementations are manually designed. In this paper, we describe a simple procedure called \textbf{AutoAugment} to automatically search for improved data augmentation policies.
In our implementation, we have designed a search space where a policy consists of many sub-policies, one of which is randomly chosen for each image in each mini-batch. A sub-policy consists of two operations, each operation being an image processing function such as translation, rotation, or shearing, and the probabilities and magnitudes with which the functions are applied. 
We use a search algorithm to find the best policy such that the neural network yields the highest validation accuracy on a target dataset. Our method achieves state-of-the-art accuracy on CIFAR-10, CIFAR-100, SVHN, and ImageNet (without additional data). On ImageNet, we attain a Top-1 accuracy of 83.5\% which is 0.4\% better than the previous record of 83.1\%. On CIFAR-10, we achieve an error rate of 1.5\%, which is 0.6\% better than the previous state-of-the-art. Augmentation policies we find are  transferable between datasets.  The policy learned on ImageNet transfers well to achieve significant improvements on other datasets, such as Oxford Flowers, Caltech-101, Oxford-IIT Pets, FGVC Aircraft, and Stanford Cars. 


\end{abstract}

\section{Introduction}

Deep neural nets are powerful machine learning systems that tend to work well when trained on massive amounts of data.  Data augmentation is an effective technique to increase both the amount and diversity of data by randomly ``augmenting" it~\cite{baird1992document,simard2003best,krizhevsky2012imagenet}; in the image domain, common augmentations include translating the image by a few pixels, or flipping the image horizontally.  Intuitively, data augmentation is used to teach a model about invariances in the data domain: classifying an object is often insensitive to horizontal flips or translation.
Network architectures can also be used to hardcode invariances: convolutional networks bake in translation invariance~\cite{fukushima1982neocognitron,lecun1998gradient,jarrett2009best,krizhevsky2012imagenet}.
However, using data augmentation to incorporate potential invariances can be easier than hardcoding invariances into the model architecture directly.

\begin{table}[h!]
\centering
\small
\begin{tabular}{llrr}
  \thickhline
   Dataset &  GPU & Best published & Our results \\
   & hours &  results & \\
  \hline 
  CIFAR-10  & 5000 & 2.1  & 1.5  \\ 
  CIFAR-100 & 0 & 12.2  & 10.7  \\ 
  SVHN & 1000& 1.3  & 1.0  \\ 
  Stanford Cars &0&5.9 & 5.2  \\ 
  ImageNet & 15000& 3.9  & 3.5  \\ 
  \thickhline
\end{tabular}
\caption{Error rates (\%) from this paper compared to the best results so far on five datasets (Top-5 for ImageNet, Top-1 for the others). Previous best result on Stanford Cars fine-tuned weights originally trained on a larger dataset~\cite{yu2017deep}, whereas we use a randomly initialized network. Previous best results on other datasets only include models that were not trained on additional data, for a single evaluation (without ensembling). See Tables~\ref{tab:small_results},\ref{tab:imagenet_results}, and \ref{tab:fgvc_results} for more detailed comparison. GPU hours are estimated for an NVIDIA Tesla P100.
}
\label{tab:summary_table}  
\end{table}

Yet a large focus of the machine learning and computer vision community has been to engineer better network architectures (e.g.,~\cite{simonyan2014very,szegedy2015going,he2016deep,szegedy2017inception,xie2017aggregated,han2017deep,zoph2017learning,hu2017squeeze,real2018regularized}). Less attention has been paid to finding better data augmentation methods that incorporate more invariances. For instance, on ImageNet, the data augmentation approach by~\cite{krizhevsky2012imagenet}, introduced in 2012, remains the standard with small changes. Even when augmentation improvements have been found for a particular dataset, they often do not transfer to other datasets as effectively. For example, horizontal flipping of images during training is an effective data augmentation method on CIFAR-10, but not on MNIST, due to the different symmetries present in these datasets. The need for automatically learned data-augmentation has been raised recently as an important unsolved problem~\cite{openai-rfr2}.

In this paper, we aim to automate the process of finding an effective data augmentation policy for a target dataset. In our implementation (Section \ref{sec:AutoAugment}), each policy expresses several choices and orders of possible augmentation operations, where each operation is an image processing function (e.g., translation, rotation, or color normalization), the probabilities of applying the function, and the magnitudes with which they are applied. We use a search algorithm to find the best choices and orders of these operations such that training a neural network yields the best validation accuracy. In our experiments, we use Reinforcement Learning~\cite{zoph2016neural} as the search algorithm, but we believe the results can be further improved if better algorithms are used~\cite{real2018regularized,mania2018simple}.

Our extensive experiments show that AutoAugment achieves excellent improvements in two use cases: 1) AutoAugment can be applied directly on the dataset of interest to  find the best augmentation policy (AutoAugment-direct) and 2) learned policies can be transferred to new datasets (AutoAugment-transfer). Firstly, for direct application, our method achieves state-of-the-art accuracy on datasets such as CIFAR-10, reduced CIFAR-10, CIFAR-100, SVHN, reduced SVHN, and ImageNet (without additional data). On CIFAR-10, we achieve an error rate of 1.5\%, which is 0.6\% better than the previous state-of-the-art~\cite{real2018regularized}. On SVHN, we improve the state-of-the-art error rate from 1.3\%~\cite{cutout2017} to 1.0\%. On reduced datasets, our method achieves performance comparable to semi-supervised methods without using any unlabeled data. On ImageNet, we achieve a Top-1 accuracy of 83.5\% which is 0.4\% better than the previous record of 83.1\%. Secondly, if direct application is too expensive, transferring an augmentation policy can be a good alternative. For transferring an augmentation policy, we show that policies found on one task can generalize well across different models and datasets. For example, the policy found on ImageNet leads to significant improvements on a variety of FGVC datasets. Even on datasets for which fine-tuning weights pre-trained on ImageNet does not help significantly~\cite{kornblith2018do}, e.g. Stanford Cars~\cite{krause2013collecting} and FGVC Aircraft~\cite{maji2013fine}, training with the ImageNet policy reduces test set error by 1.2\% and 1.8\%, respectively. This result suggests that transferring data augmentation policies offers an alternative method for standard weight transfer learning. A summary of our results is shown in Table~\ref{tab:summary_table}.

\section{Related Work}
Common data augmentation methods for image recognition have been designed manually and the best augmentation strategies are dataset-specific. For example, on MNIST, most top-ranked models use elastic distortions, scale, translation, and rotation~\cite{simard2003best,ciregan2012multi,wan2013regularization,sato2015apac}. On natural image datasets, such as CIFAR-10 and ImageNet, random cropping, image mirroring and color shifting / whitening are more common~\cite{krizhevsky2012imagenet}. As these methods are designed manually, they require expert knowledge and time.  Our approach of learning data augmentation policies from data in principle can be used for any dataset, not just one.

This paper introduces an automated approach to find data augmentation policies from data. Our approach is inspired by recent advances in architecture search, where reinforcement learning and evolution have been used to discover model architectures from data~\cite{zoph2016neural,baker2016designing,zoph2017learning,brock2017smash,liu2017hierarchical,elsken2017simple,liu2017progressive,pham2018efficient,real2017large,xie2017genetic,real2018regularized, cubuk2017intriguing}. Although these methods have improved upon human-designed architectures, it has not been possible to beat the 2\% error-rate barrier on CIFAR-10 using architecture search alone. 

Previous attempts at learned data augmentations include Smart Augmentation, which proposed a network that automatically generates augmented data by merging two or more samples from the same class~\cite{lemley2017smart}. Tran et al. used a Bayesian approach to generate data based on the distribution learned from the training set~\cite{tran2017bayesian}. DeVries and Taylor used simple transformations in the learned feature space to augment data~\cite{devries2017dataset}. 

Generative adversarial networks have also been used for the purpose of generating additional data (e.g.,~\cite{perez2017effectiveness,mun2017generative,zhu2017data,antoniou2017data,sixt2016rendergan}). The key difference between our method and generative models is that our method generates symbolic transformation operations, whereas generative models, such as GANs, generate the augmented data directly. An exception is work by Ratner et al., who used GANs to generate sequences that describe data augmentation strategies~\cite{ratner2017learning}. 

\section{AutoAugment: Searching for best Augmentation policies Directly on the Dataset of Interest}
\label{sec:AutoAugment}

\begin{figure}[h!]
\centering
\includegraphics[width=0.8\linewidth]{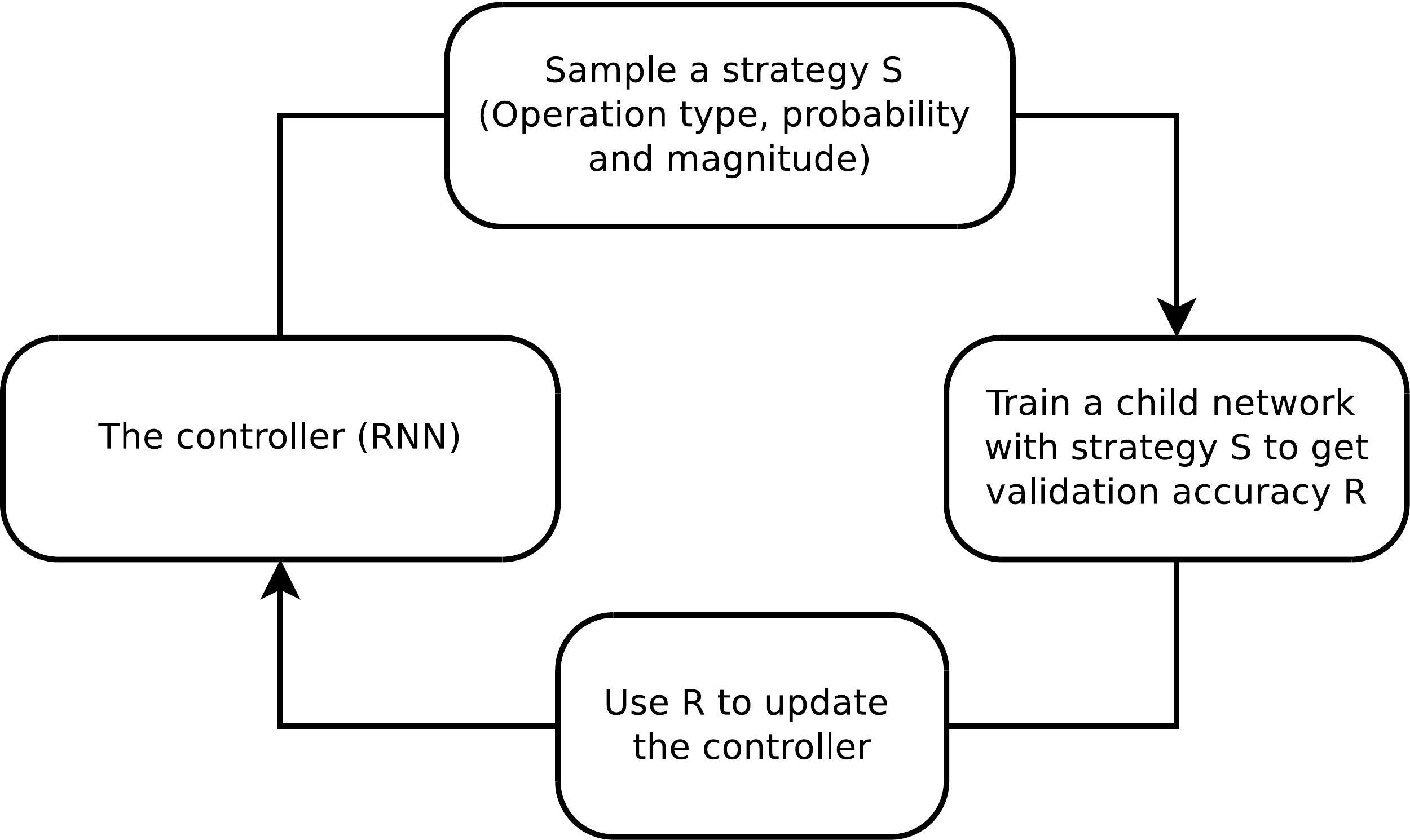}
\caption{Overview of our framework of using a search method (e.g., Reinforcement Learning) to search for better data augmentation policies. A controller
RNN predicts an augmentation policy from the search space. A child network with a fixed architecture is trained to convergence
achieving accuracy R. The reward R will be used with the policy gradient method to update the controller so that it can generate better policies over time.}
\label{fig:strategy_search}
\end{figure}

We formulate the problem of finding the best augmentation policy as a discrete search problem (see Figure~\ref{fig:strategy_search}). Our method consists of two components: A search algorithm and a search space. At a high level, the search algorithm (implemented as a controller RNN) samples a data augmentation policy $S$, which has information about what image processing operation to use, the probability of using the operation in each batch, and the magnitude of the operation. Key to our method is the fact that the policy $S$ will be used to train a neural network with a fixed architecture, whose validation accuracy $R$ will be sent back to update the controller. Since $R$ is not differentiable, the controller will be updated by policy gradient methods. In the following section we will describe the two components in detail.

\textbf{Search space details:} In our search space, a policy consists of 5 sub-policies with each sub-policy consisting of two image operations to be applied in sequence. Additionally, each operation is also associated with two hyperparameters: 1) the probability of applying the operation, and 2) the magnitude of the operation.

\begin{figure}[h!]
\centering
\includegraphics[width=1.0\linewidth]{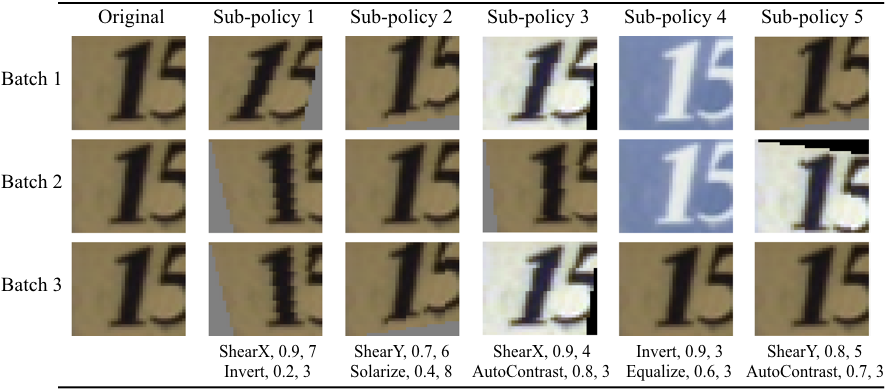}
\caption{One of the policies found on SVHN, and how it can be used to generate augmented data given an original image used to train a neural network. The policy has 5 sub-policies. For every image in a mini-batch, we choose a sub-policy uniformly at random to generate a transformed image to train the neural network. Each sub-policy consists of 2 operations, each operation is associated with two numerical values: the probability of calling the operation, and the magnitude of the operation. There is a probability of calling an operation, so the operation may not be applied in that mini-batch. However, if applied, it is applied with the fixed magnitude. We highlight the stochasticity in applying the sub-policies by showing how one image can be transformed differently in different mini-batches, even with the same sub-policy. As explained in the text, on SVHN, geometric transformations are picked more often by AutoAugment. It can be seen why Invert is a commonly selected operation on SVHN, since the numbers in the image are invariant to that transformation.}
\label{fig:strategy_viz2}
\end{figure}

Figure~\ref{fig:strategy_viz2} shows an example of a policy with 5-sub-policies in our search space. The first sub-policy specifies a sequential application of ShearX followed by Invert. The probability of applying ShearX is 0.9, and when applied, has a magnitude of 7 out of 10.  We then apply Invert with probability of 0.8. The Invert operation does not use the magnitude information. We emphasize that these operations are applied in the specified order.

 The operations we used in our experiments are from PIL, a popular Python image library.\footnote{\url{https://pillow.readthedocs.io/en/5.1.x/}} For generality, we considered all functions in PIL that accept an image as input and output an image. We additionally used two other promising augmentation techniques: Cutout~\cite{cutout2017} and SamplePairing~\cite{inoue2018data}. The operations we searched over are ShearX/Y, TranslateX/Y, Rotate, AutoContrast, Invert, Equalize, Solarize, Posterize, Contrast, Color, Brightness, Sharpness, Cutout~\cite{cutout2017}, Sample Pairing~\cite{inoue2018data}.\footnote{Details about these operations are listed in Table 1 in the Appendix.} In total, we have 16 operations in our search space. Each operation also comes with a default range of magnitudes, which will be described in more detail in Section~\ref{sec:Experiments}. We discretize the range of magnitudes into 10 values (uniform spacing) so that we can use a discrete search algorithm to find them. Similarly, we also discretize the probability of applying that operation into 11 values (uniform spacing). Finding each sub-policy becomes a search problem in a space of $(16 \times 10 \times 11)^2$ possibilities. Our goal, however, is to find 5 such sub-policies concurrently in order to increase diversity. The search space with 5 sub-policies then has roughly $(16 \times 10 \times 11)^{10}\approx 2.9\times 10^{32}$ possibilities.

The 16 operations we used and their default range of values are shown in Table 1 in the Appendix.
Notice that there is no explicit ``Identity" operation in our search space; this operation is implicit, and can be achieved by calling an operation with probability set to be 0.

\textbf{Search algorithm details: } The search algorithm that we used in our experiment uses Reinforcement Learning, inspired by~\cite{zoph2016neural,baker2016designing,zoph2017learning,bello2017neural}. The search algorithm has two components: a controller, which is a recurrent neural network, and the training algorithm, which is the Proximal Policy Optimization algorithm~\cite{schulman2017proximal}. At each step, the controller predicts a decision produced by a softmax; the prediction is then fed into the next step as an embedding. In total the controller has 30 softmax predictions in order to predict 5 sub-policies, each with 2 operations, and each operation requiring an operation type, magnitude and probability. 

\textbf{The training of controller RNN:} The controller is trained with a reward signal, which is how good the policy is in improving the generalization of a ``child model" (a neural network trained as part of the search process). In our experiments, we set aside a validation set to measure the generalization of a child model.  A child model is trained with augmented data generated by applying the 5 sub-policies on the training set (that does not contain the validation set). For each example in the mini-batch, one of the 5 sub-policies is chosen randomly to augment the image. The child model is then evaluated on the validation set to measure the accuracy, which is used as the reward signal to train the recurrent network controller. On each dataset, the controller samples about 15,000 policies. 

\textbf{Architecture of controller RNN and training hyperparameters:} We follow the training procedure and hyperparameters from \cite{zoph2017learning} for training the controller. More concretely, the controller RNN is a one-layer LSTM~\cite{hochreiter1997long} with 100
hidden units at each layer and $2 \times 5$B softmax predictions
for the two convolutional cells (where B is typically 5) associated
with each architecture decision. Each of the 10B
predictions of the controller RNN is associated with a probability.
The joint probability of a child network is the product
of all probabilities at these 10B softmaxes. This joint
probability is used to compute the gradient for the controller
RNN. The gradient is scaled by the validation accuracy of
the child network to update the controller RNN such that the
controller assigns low probabilities for bad child networks
and high probabilities for good child networks. Similar to~\cite{zoph2017learning},  we employ Proximal Policy Optimization
(PPO)~\cite{schulman2017proximal} with learning rate 0.00035. To encourage exploration
we also use an entropy penalty with a weight of
0.00001. In our implementation, the baseline function is
an exponential moving average of previous rewards with a
weight of 0.95. The weights of the controller are initialized
uniformly between -0.1 and 0.1. We choose to train the controller using PPO out of convenience, although prior work had shown that other methods (e.g. augmented random search and evolutionary strategies) can perform as well or even slightly better~\cite{kumar2018parallel}.

At the end of the search, we concatenate the sub-policies from the best 5 policies into a single policy (with 25 sub-policies). This final policy with 25 sub-policies is used to train the models for each dataset.

The above search algorithm is one of many possible search algorithms we can use to find the best policies. It might be possible to use a different discrete search algorithm such as genetic programming~\cite{real2018regularized} or even random search~\cite{bergstra2012random} to improve the results in this paper. 

\section{Experiments and Results}
\label{sec:Experiments}
\paragraph{Summary of Experiments.} In this section, we empirically investigate the performance of AutoAugment in two use cases: AutoAugment-direct and AutoAugment-transfer. First, we will benchmark AutoAugment with direct search for best augmentation policies on highly competitive datasets:  CIFAR-10~\cite{krizhevsky2009learning}, CIFAR-100~\cite{krizhevsky2009learning}, SVHN~\cite{netzer2011reading} (Section~\ref{sec:small}), and ImageNet~\cite{imagenet2009} (Section~\ref{sec:imagenet}) datasets. Our results show that a direct application of AutoAugment improves significantly the baseline models and produces state-of-the-art accuracies on these challenging datasets. Next, we will study the transferability of augmentation policies between datasets. More concretely, we will transfer the best augmentation policies found on ImageNet to fine-grained classification datasets such as Oxford 102 Flowers, Caltech-101, Oxford-IIIT Pets, FGVC Aircraft, Stanford Cars (Section~\ref{sec:transferability}). Our results also show that augmentation policies are surprisingly transferable and yield significant improvements on strong baseline models on these datasets. Finally, in Section~\ref{sec:discussion}, we will compare AutoAugment against other automated data augmentation methods and show that AutoAugment is significantly better.


\subsection{CIFAR-10, CIFAR-100, SVHN Results}
\label{sec:small}
Although CIFAR-10 has 50,000 training examples, we perform the search for the best policies on a smaller dataset we call ``reduced CIFAR-10", which consists of 4,000 randomly chosen examples, to save time for training child models during the augmentation search process (We find that the resulting policies do not seem to be sensitive to this number). We find that for a fixed amount of training time, it is more useful to allow child models to train for more epochs rather than train for fewer epochs with more training data. For the child model architecture we use small Wide-ResNet-40-2 (40 layers - widening factor of 2) model~\cite{WRN2016}, and train for 120 epochs. The use of a small Wide-ResNet is for computational efficiency as each child model is trained from scratch to compute the gradient update for the controller. We use a weight decay of 10$^{-4}$, learning rate of 0.01, and a cosine learning decay with one annealing cycle~\cite{cosine}. 

The policies found during the search on reduced CIFAR-10 are later used to train final models on CIFAR-10, reduced CIFAR-10, and CIFAR-100. As mentioned above, we concatenate sub-policies from the best 5 policies to form a single policy with 25 sub-policies, which is used for all of AutoAugment experiments on the CIFAR datasets. 

The baseline pre-processing follows the convention for state-of-the-art CIFAR-10 models: standardizing the data, using horizontal flips with 50\% probability, zero-padding and random crops, and finally Cutout with 16x16 pixels~\cite{gastaldi2017shake,yamada2018shakedrop,real2018regularized,zoph2017learning}. The AutoAugment policy is applied in addition to the standard baseline pre-processing: on one image, we first apply the baseline augmentation provided by the existing baseline methods, then apply the AutoAugment policy, then apply Cutout. We did not optimize the Cutout region size, and use the suggested value of 16 pixels~\cite{cutout2017}. Note that since Cutout is an operation in the search space, Cutout may be used twice on the same image: the first time with learned region size, and the second time with fixed region size. In practice, as the probability of the Cutout operation in the first application is small, Cutout is often used once on a given image.

On CIFAR-10, AutoAugment picks mostly color-based transformations. For example, the most commonly picked transformations on CIFAR-10 are Equalize, AutoContrast, Color, and Brightness (refer to Table 1 in the Appendix for their descriptions). Geometric transformations like ShearX and ShearY are rarely found in good policies. Furthermore, the transformation Invert is almost never applied in a successful policy. The policy found on CIFAR-10 is included in the Appendix. Below, we describe our results on the CIFAR datasets using the policy found on reduced CIFAR-10. All of the reported results are averaged over 5 runs.  

\paragraph{CIFAR-10 Results.} In Table~\ref{tab:small_results}, we show the test set accuracy on different neural network architectures. We implement the Wide-ResNet-28-10~\cite{WRN2016}, Shake-Shake~\cite{gastaldi2017shake} and ShakeDrop~\cite{yamada2018shakedrop} models in TensorFlow\cite{tensorflow-osdi2016}, and find the weight decay and learning rate hyperparameters that give the best validation set accuracy for regular training with baseline augmentation. Other hyperparameters are the same as reported in the papers introducing the models~\cite{WRN2016,gastaldi2017shake,yamada2018shakedrop}, with the exception of using a cosine learning decay for the Wide-ResNet-28-10. We then use the same model and hyperparameters to evaluate the test set accuracy of AutoAugment. For AmoebaNets, we use the same hyperparameters that were used in \cite{real2018regularized} for both baseline augmentation and AutoAugment.  As can be seen from the table, we achieve an error rate of 1.5\% with the ShakeDrop~\cite{yamada2018shakedrop} model, which is 0.6\% better than the state-of-the-art~\cite{real2018regularized}. Notice that this gain is much larger than the previous gains obtained by AmoebaNet-B against ShakeDrop (+0.2\%), and by ShakeDrop against Shake-Shake (+0.2\%). Ref.~\cite{zhang2017mixup} reports an improvement of 
1.1\% for a Wide-ResNet-28-10 model trained on CIFAR-10.
\begin{table*}[h!]
\centering
\small
\begin{tabular}{llccc}
  \thickhline
  Dataset & Model & Baseline        &  Cutout~\cite{cutout2017}    &  AutoAugment  \\
  \hline 
  \textbf{CIFAR-10} & Wide-ResNet-28-10~\cite{WRN2016} & 3.9 & 3.1 & 2.6$\pm0.1$ \\ 
  & Shake-Shake (26 2x32d)~\cite{gastaldi2017shake} & 3.6 & 3.0 & 2.5$\pm0.1$ \\
  & Shake-Shake (26 2x96d)~\cite{gastaldi2017shake}& 2.9 & 2.6 & 2.0$\pm0.1$ \\
  & Shake-Shake (26 2x112d)~\cite{gastaldi2017shake} & 2.8  & 2.6 & 1.9$\pm0.1$ \\
  & AmoebaNet-B (6,128)~\cite{real2018regularized} & 3.0 & 2.1& 1.8$\pm0.1$ \\ 
  & PyramidNet+ShakeDrop~\cite{yamada2018shakedrop}  & 2.7 & 2.3 & $\mathbf{1.5\pm0.1}$  \\ 
  \hline
  \textbf{Reduced CIFAR-10} & Wide-ResNet-28-10~\cite{WRN2016} & 18.8 & 16.5 & 14.1$\pm0.3$ \\
  & Shake-Shake (26 2x96d)~\cite{gastaldi2017shake}& 17.1 & 13.4 & $\mathbf{10.0\pm0.2}$ \\ 
  \hline
  \textbf{CIFAR-100} & Wide-ResNet-28-10~\cite{WRN2016} & 18.8 & 18.4 & 17.1$\pm0.3$ \\ 
  & Shake-Shake (26 2x96d)~\cite{gastaldi2017shake}& 17.1 & 16.0 & 14.3$\pm0.2$ \\
  & PyramidNet+ShakeDrop~\cite{yamada2018shakedrop}  & 14.0 & 12.2 & $\mathbf{10.7\pm0.2}$  \\
  \hline
  \textbf{SVHN} & Wide-ResNet-28-10~\cite{WRN2016} & 1.5 & 1.3 & 1.1    \\
  & Shake-Shake (26 2x96d)~\cite{gastaldi2017shake} & 1.4 & 1.2 & \textbf{1.0}   \\ 
  \hline
  \textbf{Reduced SVHN} & Wide-ResNet-28-10~\cite{WRN2016} & 13.2 & 32.5 & 8.2    \\
  & Shake-Shake (26 2x96d)~\cite{gastaldi2017shake} & 12.3 &24.2 & \textbf{5.9}   \\ 
  \thickhline
\end{tabular}
\caption{Test set error rates (\%) on CIFAR-10, CIFAR-100, and SVHN datasets. Lower is better. All the results of the baseline models, and baseline models with Cutout are replicated in our experiments and match the previously reported results~\cite{WRN2016,gastaldi2017shake,yamada2018shakedrop,cutout2017}. Two exceptions are Shake-Shake (26 2x112d), which has more filters than the biggest model in~\cite{gastaldi2017shake} -- 112 vs 96, and Shake-Shake models trained on SVHN, these results were not previously reported. See text for more details.}
\label{tab:small_results}  
\end{table*}

We also evaluate our best model trained with AutoAugment on a recently proposed CIFAR-10 test set~\cite{recht2018cifar}. Recht et al.~\cite{recht2018cifar} report that Shake-Shake (26 2x64d) + Cutout performs best on this new dataset, with an error rate of 7.0\% (4.1\% higher relative to error rate on the original CIFAR-10 test set). Furthermore, PyramidNet+ShakeDrop achieves an error rate of 7.7\% on the new dataset (4.6\% higher relative to the original test set). Our best model, PyramidNet+ShakeDrop trained with AutoAugment achieves an error rate of 4.4\% (2.9\% higher than the error rate on the original set). Compared to other models evaluated on this new dataset, our model exhibits a significantly smaller drop in accuracy.

\paragraph{CIFAR-100 Results.} We also train models on CIFAR-100 with the same AutoAugment policy found on reduced-CIFAR-10; results are shown in Table~\ref{tab:small_results}. Again, we achieve the state-of-art result on this dataset, beating the previous record of 12.19\% error rate by ShakeDrop regularization~\cite{yamada2018shakedrop}.

Finally, we apply the same AutoAugment policy to train models on reduced CIFAR-10 (the same 4,000 example training set that we use to find the best policy). Similar to the experimental convention used by the semi-supervised learning community~\cite{tarvainen2017mean,miyato2017virtual,sajjadi2016regularization,laine2016temporal,oliver2018realistic} we train on 4,000 labeled samples. But we do not use the 46,000 unlabeled samples during training. Our results shown in Table~\ref{tab:small_results}. We note that the improvement in accuracy due to AutoAugment is more significant on the reduced dataset compared to the full dataset. As the size of the training set grows, we expect that the effect of data-augmentation will be reduced. However, in the next sections we show that even for larger datasets like SVHN and ImageNet, AutoAugment can still lead to improvements in generalization accuracy.  

\paragraph{SVHN Results}
We experimented with the SVHN dataset~\cite{netzer2011reading}, which has 73,257 training examples (also called ``core training set"), and 531,131 additional training examples. The test set has 26,032 examples. To save time during the search, we created a reduced SVHN dataset of 1,000 examples sampled randomly from the core training set.  We use AutoAugment to find the best policies. The model architecture and training procedure of the child models are identical to the above experiments with CIFAR-10.  

The policies picked on SVHN are different than the transformations picked on CIFAR-10. For example, the most commonly picked transformations on SVHN are Invert, Equalize, ShearX/Y, and Rotate. As mentioned above, the transformation Invert is almost never used on CIFAR-10, yet it is very common in successful SVHN policies. Intuitively, this makes sense since the specific color of numbers is not as important as the relative color of the number and its background. Furthermore, geometric transformations ShearX/Y are two of the most popular transformations on SVHN. This also can be understood by general properties of images in SVHN: house numbers are often naturally sheared and skewed in the dataset, so it is helpful to learn the invariance to such transformations via data augmentation. Five successful sub-policies are visualized on SVHN examples in Figure~\ref{fig:strategy_viz2}.

After the end of the search, we concatenate the 5 best policies and apply them to train architectures that already perform well on SVHN using standard augmentation policies. For full training, we follow the common procedure mentioned in the Wide-ResNet paper~\cite{WRN2016} of using the core training set and the extra data. The validation set is constructed by setting aside the last 7325 samples of the training set. We tune the weight decay and learning rate on the validation set performance. Other hyperparameters and training details are identical to the those in the papers introducing the models~\cite{WRN2016,gastaldi2017shake}. One exception is that we trained the Shake-Shake model only for 160 epochs (as opposed to 1,800), due to the large size of the full SVHN dataset. Baseline pre-processing involves standardizing the data and applying Cutout with a region size of 20x20 pixels, following the procedure outlined in \cite{cutout2017}. AutoAugment results combine the baseline pre-processing with the policy learned on SVHN. One exception is that we do not use Cutout on reduced SVHN as it lowers the accuracy significantly. The summary of the results in this experiment are shown in Table~\ref{tab:small_results}. As can be seen from the table, we achieve state-of-the-art accuracy using both models.

We also test the best policies on reduced SVHN (the same 1,000 example training set where the best policies are found). AutoAugment results on the reduced set are again comparable to the leading semi-supervised methods, which range from 5.42\% to 3.86\%~\cite{miyato2017virtual}. (see Table~\ref{tab:small_results}). We see again that AutoAugment leads to more significant improvements on the reduced dataset than the full dataset.

\subsection{ImageNet Results}
\label{sec:imagenet}
\begin{figure}[h!]
\centering
\includegraphics[width=1.0\linewidth]{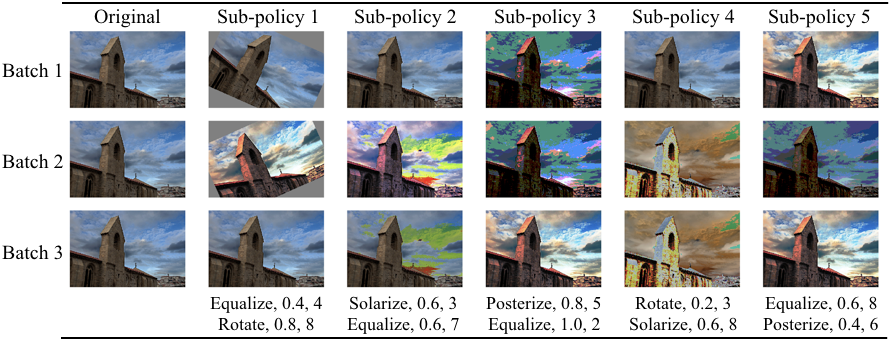}
\caption{One of the successful policies on ImageNet. As described in the text, most of the policies found on ImageNet used color-based transformations.
}
\label{fig:strategy_viz1}
\end{figure}

Similar to above experiments, we use a reduced subset of the ImageNet training set, with 120 classes (randomly chosen) and 6,000 samples, to search for policies. We train a Wide-ResNet 40-2 using cosine decay for 200 epochs. A weight decay of $10^{-5}$ was used along with a learning rate of 0.1. 
The best policies found on ImageNet are similar to those found on CIFAR-10, focusing on color-based transformations. One difference is that a geometric transformation, Rotate, is commonly used on ImageNet policies. One of the best policies is visualized in Figure~\ref{fig:strategy_viz1}. 

Again, we combine the 5 best policies for a total of 25 sub-policies to create the final policy for ImageNet training. We then train on the full ImageNet from scratch with this policy using the ResNet-50 and ResNet-200 models for 270 epochs. We use a batch size of 4096 and a learning rate of 1.6. We decay the learning rate by 10-fold at epochs 90, 180, and 240. For baseline augmentation, we use the standard Inception-style pre-processing which involves scaling pixel values to [-1,1], horizontal flips with 50\% probability, and random distortions of colors~\cite{howard2013some,szegedy2015going}. For models trained with AutoAugment, we use the baseline pre-processing and the policy learned on ImageNet. We find that removing the random distortions of color does not change the results for AutoAugment.    
\begin{table}[h!]
\centering
\small
\begin{tabular}{lcc}
    \thickhline
  Model   &  Inception     &  AutoAugment \\  
  & Pre-processing~\cite{szegedy2015going} & ours \\
  \hline 
  ResNet-50  & 76.3 / 93.1 & 77.6 / 93.8  \\ 
  ResNet-200 &  78.5 / 94.2 & 80.0 / 95.0 \\ 
  AmoebaNet-B (6,190)  & 82.2 / 96.0 & 82.8 / 96.2 \\
  AmoebaNet-C (6,228)  & 83.1 / 96.1 & \textbf{83.5 / 96.5} \\
  \thickhline
\end{tabular}
\caption{Validation set Top-1 / Top-5 accuracy (\%) on ImageNet.  Higher is better. ResNet-50 with baseline augmentation result is taken from \cite{he2016deep}. AmoebaNet-B,C results with Inception-style preprocessing are replicated in our experiments and match the previously reported result by \cite{real2018regularized}. There exists a better result of 85.4\% Top-1 error rate~\cite{fb2018} but their method makes use of a large amount of weakly labeled extra data. Ref.~\cite{zhang2017mixup} reports an improvement of 1.5\% for a ResNet-50 model.
}
\label{tab:imagenet_results}  
\end{table}

Our ImageNet results are shown in Table~\ref{tab:imagenet_results}. As can be seen from the results, AutoAugment improves over the widely-used Inception Pre-processing~\cite{szegedy2015going} across a wide range of models, from ResNet-50 to the state-of-art AmoebaNets~\cite{real2018regularized}. Secondly, applying AutoAugment to AmoebaNet-C improves its top-1 and top-5 accuracy from 83.1\% / 96.1\% to 83.5\% / 96.5\%. This improvement is remarkable given that the best augmentation policy was discovered on  5,000 images. We expect the results to be even better when more compute is available so that AutoAugment can use more images to discover even better augmentation policies. The accuracy of 83.5\% / 96.5\% is also the new state-of-art top-1/top-5 accuracy on this dataset (without multicrop / ensembling).

\subsection{The Transferability of Learned Augmentation policies to Other Datasets} 
\label{sec:transferability}

In the above, we applied AutoAugment directly to find augmentation policies on the dataset of interest (AutoAugment-direct). In many cases, such application of AutoAugment can be resource-intensive. Here we seek to understand if it is possible to transfer augmentation policies from one dataset to another (which we call AutoAugment-transfer). If such transfer happens naturally, the resource requirements won't be as intensive as applying AutoAugment directly. Also if such transfer happens naturally, we also have clear evidence that AutoAugment does not ``overfit'' to the dataset of interest and that AutoAugment indeed finds generic transformations that can be applied to all kinds of problems.

To evaluate the transferability of the policy found on ImageNet, we use the same policy that is learned on ImageNet (and used for the results on Table~\ref{tab:imagenet_results}) on five FGVC datasets with image size similar to ImageNet. These datasets are challenging as they have relatively small sets of training examples while having a large number of classes.

\begin{table}[h!]
\centering
\footnotesize
\begin{tabular}{lcccc}
  \thickhline
  Dataset & Train  & Classes &  Baseline    &  AutoAugment- \\
          & Size   &         &        & transfer  \\
  \hline
  Oxford 102  & 2,040 & 102 & 6.7 & \bf{4.6}  \\
  Flowers~\cite{nilsback2008automated} & & & & \\ \hline
  Caltech-101~\cite{fei2007learning} & 3,060 & 102 & 19.4 & \bf{13.1} \\ \hline
  Oxford-IIIT  & 3,680 & 37 & 13.5 & \bf{11.0} \\
  Pets~\cite{em2017incorporating} & & & & \\ \hline
  FGVC   & 6,667 & 100& 9.1 & \bf{7.3} \\ 
  Aircraft~\cite{maji2013fine}  & & & & \\ \hline
  Stanford   & 8,144 & 196 & 6.4 & \bf{5.2} \\
  Cars~\cite{krause2013collecting} & & & & \\
  \thickhline
\end{tabular}
\caption{Test set Top-1 error rates (\%) on FGVC datasets for Inception v4 models trained from scratch with and without AutoAugment-transfer. Lower rates are better. AutoAugment-transfer results use the policy found on ImageNet. Baseline models used Inception pre-processing.}
\label{tab:fgvc_results}  
\end{table}

For all of the datasets listed in Table~\ref{tab:fgvc_results}, we train a Inception v4~\cite{szegedy2017inception} for 1,000 epochs, using a cosine learning rate decay with one annealing cycle. The learning rate and weight decay are chosen based on the validation set performance. We then combine the training set and the validation set and train again with the chosen hyperparameters. The image size is set to 448x448 pixels. The policies found on ImageNet improve the generalization accuracy of all of the FGVC datasets significantly. To the best of our knowledge, our result on the Stanford Cars dataset is the lowest error rate achieved on this dataset although we train the network weights from scratch. Previous state-of-the-art fine-tuned pre-trained weights on ImageNet and used deep layer aggregation to attain a 5.9\% error rate~\cite{yu2017deep}.

\section{Discussion}
\label{sec:discussion}

In this section, we compare our search to previous attempts at automated data augmentation methods. We also discuss the dependence of our results on some of the design decisions we have made through several ablation experiments.

\textbf{AutoAugment vs. other automated data augmentation methods:} Most notable amongst many previous data augmentation methods is the work of~\cite{ratner2017learning}. The setup in~\cite{ratner2017learning} is similar to GANs~\cite{goodfellow2014generative}: a generator learns to propose augmentation policy (a sequence of image processing operations) such that the augmented images can fool a discriminator. The difference of our method to theirs is that our method tries to optimize classification accuracy directly whereas their method just tries to make sure the augmented images are similar to the current training images. 

To make the comparison fair, we carried out experiments similar to that described in~\cite{ratner2017learning}. We trained a ResNet-32 and a ResNet-56 using the same policy from Section~\ref{sec:small}, to compare our method to the results from~\cite{ratner2017learning}. By training a ResNet-32 with Baseline data augmentation, we achieve the same error as~\cite{ratner2017learning} did with ResNet-56 (called Heur. in~\cite{ratner2017learning}). For this reason, we trained both a ResNet-32 and a ResNet-56. We show that for both models, AutoAugment leads to higher improvement ($\sim$3.0\%). 
\begin{table}[h!]
\centering
\small
\begin{tabular}{lccc}
  \hline
  Method       & Baseline    &  Augmented & Improvement $\Delta$  \\
  \hline
  LSTM~\cite{ratner2017learning}  &  7.7 & 6.0 & 1.6 \\ 
  \hline
  MF~\cite{ratner2017learning}  & 7.7 & 5.6 & 2.1 \\ 
    \hline
  AutoAugment & 7.7 & 4.5 & \textbf{3.2} \\
  (ResNet-32) & & \\
    \hline
  AutoAugment &  6.6 & 3.6 & \textbf{3.0} \\
  (ResNet-56) & & \\
    \hline
\end{tabular}
\caption{The test set error rates ($\%$) on CIFAR-10 with different approaches for automated data augmentation. The MF and LSTM results are taken from~\cite{ratner2017learning}, and they are for a ResNet-56.}
\end{table}

\textbf{Relation between training steps and number of sub-policies:} An important aspect of our work is the stochastic application of sub-policies during training. Every image is only augmented by one of the many sub-policies available in each mini-batch, which itself has further stochasticity since each transformation has a probability of application associated with it. We find that this stochasticity requires a certain number of epochs per sub-policy for AutoAugment to be effective. Since the child models are each trained with 5 sub-policies, they need to be trained for more than 80-100 epochs before the model can fully benefit from all of the sub-policies. This is the reason we choose to train our child models for 120 epochs. Each sub-policy needs to be applied a certain number of times before the model benefits from it. After the policy is learned, the full model is trained for longer (e.g. 1800 epochs for Shake-Shake on CIFAR-10, and 270 epochs for ResNet-50 on ImageNet), which allows us to use more sub-policies.

\textbf{Transferability across datasets and architectures:} It is important to note that the policies described above transfer well to many model architectures and datasets. For example, the policy learned on Wide-ResNet-40-2 and reduced CIFAR-10 leads to the improvements described on all of the other model architectures trained on full CIFAR-10 and CIFAR-100. Similarly, a policy learned on Wide-ResNet-40-2 and reduced ImageNet leads to significant improvements on Inception v4 trained on FGVC datasets that have different data and class distributions. AutoAugment policies are never found to hurt the performance of models even if they are learned on a different dataset, which is not the case for Cutout on reduced SVHN (Table~\ref{tab:small_results}). We present the best policy on ImageNet and SVHN in the Appendix, which can hopefully help researchers improve their generalization accuracy on relevant image classification tasks. 

Despite the observed transferability, we find that policies learned on data distributions closest to the target yield the best performance: when training on SVHN, using the best policy learned on reduced CIFAR-10 does slightly improve generalization accuracy compared to the baseline augmentation, but not as significantly as applying the SVHN-learned policy.

\subsection{Ablation experiments}
\begin{figure}[h!]
\centering
\includegraphics[width=0.7\linewidth]{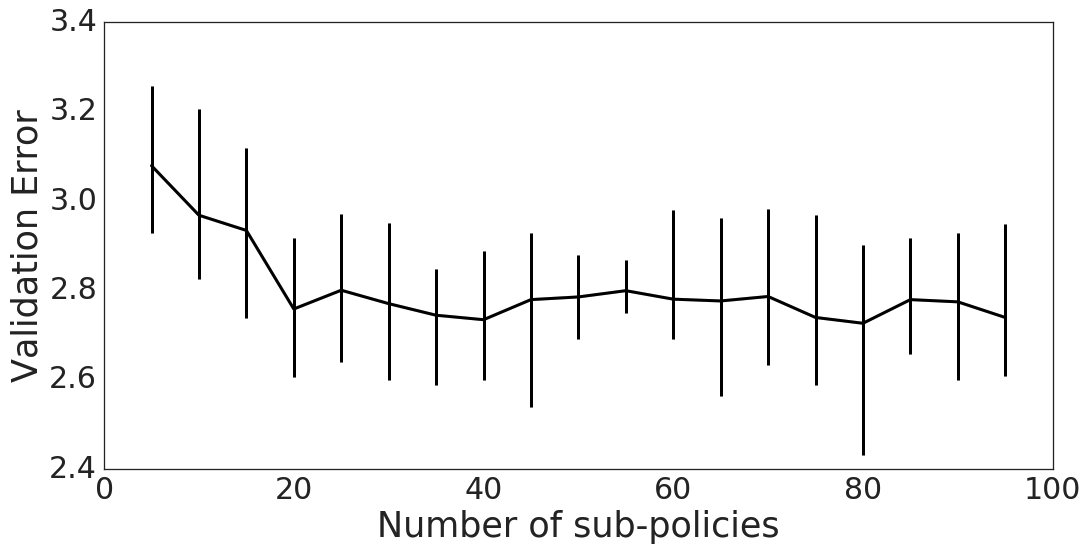}
\caption{Validation error (averaged over 5 runs) of Wide-ResNet-28-10 trained on CIFAR-10 as a function of number of \textit{randomly selected} sub-policies (out of a pool of 500 good sub-policies) used in training with AutoAugment. Bars represent the range of validation errors for each number.}
\label{fig:acc_vs_no_pol}
\end{figure}

\textbf{Changing the number of sub-policies:} Our hypothesis is that as we increase the number of sub-policies, the neural network is trained on the same points with a greater diversity of augmentation, which should increase the generalization accuracy. To test this hypothesis, we investigate the average validation accuracy of fully-trained Wide-ResNet-28-10 models on CIFAR-10 as a function of the number of sub-policies used in training. We randomly select sub-policy sets from a pool of 500 good sub-policies, and train the Wide-ResNet-28-10 model for 200 epochs with each of these sub-policy sets. For each set size, we sampled sub-policies five different times for better statistics. The training details of the model are the same as above for Wide-ResNet-28-10 trained on CIFAR-10. Figure~\ref{fig:acc_vs_no_pol} shows the average validation set accuracy as a function of the number of sub-policies used in training, confirming that the validation accuracy improves with more sub-policies up to about 20 sub-policies.

\textbf{Randomizing the probabilities and magnitudes in the augmentation policy:} We take the AutoAugment policy on CIFAR-10 and randomize the probabilities and magnitudes of each operation in it. We train a Wide-ResNet-28-10~\cite{WRN2016}, using the same training procedure as before, for 20 different instances of the randomized probabilities and magnitudes. We find the average error to be 3.0\% (with a standard deviation of 0.1\%), which is 0.4\% worse than the result achieved with the original AutoAugment policy (see Table~\ref{tab:small_results}). 

\textbf{Performance of random policies:} Next, we randomize the whole policy, the operations as well as the probabilities and magnitudes. Averaged over 20 runs, this experiment yields an average accuracy of 3.1\% (with a standard deviation of 0.1\%), which is slightly worse than randomizing only the probabilities and magnitudes. The best random policy achieves achieves an error of 3.0\% (when average over 5 independent runs). This shows that even AutoAugment with randomly sampled policy leads to appreciable improvements. 

The ablation experiments indicate that even data augmentation policies that are randomly sampled from our search space can lead to improvements on CIFAR-10 over the baseline augmentation policy. However, the improvements exhibited by random policies are less than those shown by the AutoAugment policy ($2.6\%\pm0.1\%$ vs. $3.0\%\pm0.1\%$ error rate). Furthermore, the probability and magnitude information learned within the AutoAugment policy seem to be important, as its effectiveness is reduced significantly when those parameters are randomized. We emphasize again that we trained our controller using RL out of convenience, augmented random search and evolutionary strategies can be used just as well. The main contribution of this paper is in our approach to data augmentation and in the construction of the search space; not in discrete optimization methodology.

\section{Acknowledgments}

We thank Alok Aggarwal, Gabriel Bender, Yanping Huang, Pieter-Jan Kindermans, Simon Kornblith, Augustus Odena, Avital Oliver, Colin Raffel, and Jonathan Shlens for helpful discussions. This
work was done as part of the Google Brain Residency program (g.co/brainresidency).

{\small
\bibliographystyle{ieee}

}

\clearpage
\appendix


\section{Supplementary materials for ``AutoAugment: Learning Augmentation policies from Data"}

\begin{table*}[tbh]
\centering
\small
\begin{tabular}{l p{8cm} p{1.35cm}}
\thickhline
  Operation Name & Description & Range of \\ 
                 &             & magnitudes \\ 
  \hline
  ShearX(Y) & Shear the image along the horizontal (vertical) axis with rate \emph{magnitude}.  & [-0.3,0.3]  \\ 
  TranslateX(Y) & Translate the image in the horizontal (vertical) direction by \emph{magnitude} number of pixels.   & [-150,150] \\ 
  Rotate & Rotate the image \emph{magnitude} degrees. & [-30,30] \\ 
  AutoContrast & Maximize the the image contrast, by making the darkest pixel black and lightest pixel white.  &\\ 
  Invert & Invert the pixels of the image. &\\ 
  Equalize & Equalize the image histogram. &\\ 
  Solarize & Invert all pixels above a threshold value of \emph{magnitude}.  & [0,256] \\ 
  Posterize & Reduce the number of bits for each pixel to \emph{magnitude} bits.  & [4,8]\\ 
  Contrast & Control the contrast of the image. A \emph{magnitude}=0 gives a gray image, whereas \emph{magnitude}=1 gives the original image.    & [0.1,1.9]\\
  Color & Adjust the color balance of the image,  in a manner similar to the controls on a colour TV set. A \emph{magnitude}=0 gives a black \& white image, whereas \emph{magnitude}=1 gives the original image.  & [0.1,1.9]\\ 
  Brightness & Adjust the brightness of the image. A \emph{magnitude}=0 gives a black image, whereas \emph{magnitude}=1 gives the original image.   & [0.1,1.9] \\  
  Sharpness & Adjust the sharpness of the image. A \emph{magnitude}=0 gives a blurred image, whereas \emph{magnitude}=1 gives the original image.   & [0.1,1.9] \\ 
  Cutout~\cite{cutout2017,zhong2017random} & Set a random square patch of side-length \emph{magnitude} pixels to gray. & [0,60] \\
  Sample Pairing~\cite{inoue2018data,zhang2017mixup} & Linearly add the image with another image (selected at random from the same mini-batch) with weight \emph{magnitude}, without changing the label. & [0, 0.4] \\ 
\thickhline
\end{tabular}
\caption{List of all image transformations that the controller could choose from during the search. Additionally, the values of magnitude that can be predicted by the controller during the search for each operation at shown in the third column (for image size 331x331). Some transformations do not use the magnitude information (e.g. Invert and Equalize).}
\end{table*}

\begin{table*}[h!]
\centering
\small
\begin{tabular}{lll}
\thickhline
 & Operation 1 & Operation 2 \\
 \hline
Sub-policy 0&(Invert,0.1,7)&(Contrast,0.2,6)\\
Sub-policy 1&(Rotate,0.7,2)&(TranslateX,0.3,9)\\
Sub-policy 2&(Sharpness,0.8,1)&(Sharpness,0.9,3)\\
Sub-policy 3&(ShearY,0.5,8)&(TranslateY,0.7,9)\\
Sub-policy 4&(AutoContrast,0.5,8)&(Equalize,0.9,2)\\
Sub-policy 5&(ShearY,0.2,7)&(Posterize,0.3,7)\\
Sub-policy 6&(Color,0.4,3)&(Brightness,0.6,7)\\
Sub-policy 7&(Sharpness,0.3,9)&(Brightness,0.7,9)\\
Sub-policy 8&(Equalize,0.6,5)&(Equalize,0.5,1)\\
Sub-policy 9&(Contrast,0.6,7)&(Sharpness,0.6,5)\\
Sub-policy 10&(Color,0.7,7)&(TranslateX,0.5,8)\\
Sub-policy 11&(Equalize,0.3,7)&(AutoContrast,0.4,8)\\
Sub-policy 12&(TranslateY,0.4,3)&(Sharpness,0.2,6)\\
Sub-policy 13&(Brightness,0.9,6)&(Color,0.2,8)\\
Sub-policy 14&(Solarize,0.5,2)&(Invert,0.0,3)\\
Sub-policy 15&(Equalize,0.2,0)&(AutoContrast,0.6,0)\\
Sub-policy 16&(Equalize,0.2,8)&(Equalize,0.6,4)\\
Sub-policy 17&(Color,0.9,9)&(Equalize,0.6,6)\\
Sub-policy 18&(AutoContrast,0.8,4)&(Solarize,0.2,8)\\
Sub-policy 19&(Brightness,0.1,3)&(Color,0.7,0)\\
Sub-policy 20&(Solarize,0.4,5)&(AutoContrast,0.9,3)\\
Sub-policy 21&(TranslateY,0.9,9)&(TranslateY,0.7,9)\\
Sub-policy 22&(AutoContrast,0.9,2)&(Solarize,0.8,3)\\
Sub-policy 23&(Equalize,0.8,8)&(Invert,0.1,3)\\
Sub-policy 24&(TranslateY,0.7,9)&(AutoContrast,0.9,1)\\
\thickhline
\end{tabular}
\caption{AutoAugment policy found on reduced CIFAR-10.}

\end{table*}

\begin{table*}[h!]
\centering
\small
\begin{tabular}{lll}
\thickhline
 & Operation 1 & Operation 2 \\
 \hline
Sub-policy 0&(ShearX,0.9,4)&(Invert,0.2,3)\\
Sub-policy 1&(ShearY,0.9,8)&(Invert,0.7,5)\\
Sub-policy 2&(Equalize,0.6,5)&(Solarize,0.6,6)\\
Sub-policy 3&(Invert,0.9,3)&(Equalize,0.6,3)\\
Sub-policy 4&(Equalize,0.6,1)&(Rotate,0.9,3)\\
Sub-policy 5&(ShearX,0.9,4)&(AutoContrast,0.8,3)\\
Sub-policy 6&(ShearY,0.9,8)&(Invert,0.4,5)\\
Sub-policy 7&(ShearY,0.9,5)&(Solarize,0.2,6)\\
Sub-policy 8&(Invert,0.9,6)&(AutoContrast,0.8,1)\\
Sub-policy 9&(Equalize,0.6,3)&(Rotate,0.9,3)\\
Sub-policy 10&(ShearX,0.9,4)&(Solarize,0.3,3)\\
Sub-policy 11&(ShearY,0.8,8)&(Invert,0.7,4)\\
Sub-policy 12&(Equalize,0.9,5)&(TranslateY,0.6,6)\\
Sub-policy 13&(Invert,0.9,4)&(Equalize,0.6,7)\\
Sub-policy 14&(Contrast,0.3,3)&(Rotate,0.8,4)\\
Sub-policy 15&(Invert,0.8,5)&(TranslateY,0.0,2)\\
Sub-policy 16&(ShearY,0.7,6)&(Solarize,0.4,8)\\
Sub-policy 17&(Invert,0.6,4)&(Rotate,0.8,4)\\
Sub-policy 18&(ShearY,0.3,7)&(TranslateX,0.9,3)\\
Sub-policy 19&(ShearX,0.1,6)&(Invert,0.6,5)\\
Sub-policy 20&(Solarize,0.7,2)&(TranslateY,0.6,7)\\
Sub-policy 21&(ShearY,0.8,4)&(Invert,0.8,8)\\
Sub-policy 22&(ShearX,0.7,9)&(TranslateY,0.8,3)\\
Sub-policy 23&(ShearY,0.8,5)&(AutoContrast,0.7,3)\\
Sub-policy 24&(ShearX,0.7,2)&(Invert,0.1,5)\\
\thickhline
\end{tabular}
\caption{AutoAugment policy found on reduced SVHN.}
\end{table*}

\begin{table*}[h!]
\centering
\small
\begin{tabular}{lll}
\thickhline
 & Operation 1 & Operation 2 \\
 \hline
Sub-policy 0&(Posterize,0.4,8)&(Rotate,0.6,9)\\
Sub-policy 1&(Solarize,0.6,5)&(AutoContrast,0.6,5)\\
Sub-policy 2&(Equalize,0.8,8)&(Equalize,0.6,3)\\
Sub-policy 3&(Posterize,0.6,7)&(Posterize,0.6,6)\\
Sub-policy 4&(Equalize,0.4,7)&(Solarize,0.2,4)\\
Sub-policy 5&(Equalize,0.4,4)&(Rotate,0.8,8)\\
Sub-policy 6&(Solarize,0.6,3)&(Equalize,0.6,7)\\
Sub-policy 7&(Posterize,0.8,5)&(Equalize,1.0,2)\\
Sub-policy 8&(Rotate,0.2,3)&(Solarize,0.6,8)\\
Sub-policy 9&(Equalize,0.6,8)&(Posterize,0.4,6)\\
Sub-policy 10&(Rotate,0.8,8)&(Color,0.4,0)\\
Sub-policy 11&(Rotate,0.4,9)&(Equalize,0.6,2)\\
Sub-policy 12&(Equalize,0.0,7)&(Equalize,0.8,8)\\
Sub-policy 13&(Invert,0.6,4)&(Equalize,1.0,8)\\
Sub-policy 14&(Color,0.6,4)&(Contrast,1.0,8)\\
Sub-policy 15&(Rotate,0.8,8)&(Color,1.0,2)\\
Sub-policy 16&(Color,0.8,8)&(Solarize,0.8,7)\\
Sub-policy 17&(Sharpness,0.4,7)&(Invert,0.6,8)\\
Sub-policy 18&(ShearX,0.6,5)&(Equalize,1.0,9)\\
Sub-policy 19&(Color,0.4,0)&(Equalize,0.6,3)\\
Sub-policy 20&(Equalize,0.4,7)&(Solarize,0.2,4)\\
Sub-policy 21&(Solarize,0.6,5)&(AutoContrast,0.6,5)\\
Sub-policy 22&(Invert,0.6,4)&(Equalize,1.0,8)\\
Sub-policy 23&(Color,0.6,4)&(Contrast,1.0,8)\\
Sub-policy 24&(Equalize,0.8,8)&(Equalize,0.6,3)\\
\thickhline
\end{tabular}
\caption{AutoAugment policy found on reduced ImageNet.}
\end{table*}

\end{document}